\documentclass{article}
\usepackage{graphicx}
\usepackage{amsmath}
\usepackage{amsfonts}
\usepackage{enumitem}
\usepackage{booktabs}
\usepackage{subcaption}
\usepackage{wrapfig}
\usepackage[preprint]{corl_2026} 

\def\algorithmname{MATE}

\title{Learning Multi-Modal Trajectory Policies for Data-Efficient Robotic Manipulation}

\author{
  Zijia Chen$^{1}$ \quad
  Yuenan Hou$^{2,\dagger}$ \quad
  Xinhua Jiang$^{1}$ \quad
  Yu Li$^{2}$ \quad
  Weijie Li$^{1}$ \quad
  Li Liu$^{1,\dagger}$ \\
  $^{1}$College of Electronic Science and Technology, National University of Defense Technology  \\
  Changsha, 410073, China \\
  $^{2}$Shanghai AI Laboratory \\
  Shanghai, 200000, China \\
  {\small $^\dagger$Corresponding authors.}
}

\begin{document}
\maketitle


\begin{abstract}
    Robotic manipulation requires the effective integration of heterogeneous inputs, including visual observations, language instructions, and trajectory representations, to generate accurate actions. Existing transformer-based policies typically process these heterogeneous modalities within a shared parameter space, which often leads to modality interference and inefficient representation learning, especially in data-scarce scenarios. While Mixture-of-Experts (MoE) offers a scalable solution through expert specialization, conventional routing mechanisms are often sensitive to such cross-modal representation discrepancies, resulting in unstable expert assignment and expert collapse. In this work, we propose \textbf{MATE} (\textbf{M}ulti-Mod\textbf{A}l \textbf{T}raj\textbf{E}ctory Policies), a novel trajectory prediction framework built upon MoE. Specifically, we introduce a Multi-Modal MoE architecture to achieve fine-grained sub-token feature decoupling, and design a cross-modal cosine router for stable and scale-invariant expert assignment across heterogeneous modalities. We further employ temperature-controlled routing and stochastic noise injection to improve expert balance and prevent premature routing collapse under scarce demonstrations. Experiments on the LIBERO benchmark show that our \algorithmname~consistently outperforms prior work under data scarcity. It achieves a \textbf{4.75\%} improvement in average success rate over the trajectory-guided counterpart. Real-world experiments on robotic ping-pong also suggest that the predicted trajectories can provide useful guidance for downstream robotic execution, further indicating the practical feasibility of our algorithm.
\end{abstract}



\section{Introduction}
	
    Robotic manipulation aims to enable robots to physically interact with environments to perform complex tasks~\cite{siciliano2008springer, levine2016learning, li2025roboticmanipulationil}. In recent years, with the rapid development of deep learning, large-scale data-driven learning, and foundation models across perception and embodied-control domains~\cite{kim2025openvlaoft, pi05_2025, liu2026atrnet, cai2026internvlaa1, li2025saratr}, imitation learning has emerged as a predominant paradigm for robotic policy learning. Although imitation learning has achieved encouraging progress~\cite{levine2016end, zhang2018deep, chi2023diffusion, brohan2023rt1, driess2023palm, brohan2023rt2}, directly predicting low-level actions remains vulnerable to compounding errors in long-horizon tasks~\cite{chi2023diffusion, ross2011dagger, mandlekar2020gti, li2025clone, zhao2023learning}, while collecting diverse high-quality demonstrations is still costly.

    To address these challenges, recent research has explored learning more structured intermediate representations to improve policy generalization across different tasks and environments~\cite{mandlekar2020gti, ravichandar2020recent, chen2025generalization, bai2025embodied}. One representative direction is trajectory-guided policy learning, which reformulates manipulation policy learning as a motion trajectory modeling problem~\cite{wen2024anypoint}. By predicting structured trajectories, these methods provide motion-level guidance and reduce the reliance on expensive action-labeled demonstrations. Despite the advantages of trajectory modeling, policy learning in robotic manipulation often involves heterogeneous inputs from multiple modalities, such as visual observations, language instructions, and trajectory representations. These modalities exhibit significant differences in feature distributions and semantic structures. When processing these heterogeneous tokens, a single shared network often struggles to accommodate them simultaneously, leading to severe modality interference and optimization conflicts, which limits the model's capacity for complex manipulation. 

    To tackle this issue, recent studies have explored applying the Mixture-of-Experts architecture to trajectory prediction. For example, Tra-MoE~\cite{yang2025tramoe} dynamically allocates inputs with different semantic and distributional characteristics to specific experts, effectively mitigating cross-modal conflicts. However, the success of traditional MoE architectures relies heavily on large-scale datasets to learn and maintain balanced routing assignments~\cite{mu2025comprehensive, lepikhin2020glam, kaplan2020scaling}. In robotic manipulation scenarios where high-quality demonstration data is scarce~\cite{brohan2023rt1, liu2025rdt1b, hu2023robofm}, traditional MoE-based models are confronted with significant challenges, including \textbf{training instability and expert collapse}. Thus, how to maintain a stable and efficient training process while improving trajectory representation capability under limited data conditions remains a critical challenge.

    \begin{figure}[htbp]
        \centering
        \makebox[0pt][c]{\includegraphics[width=1\textwidth]{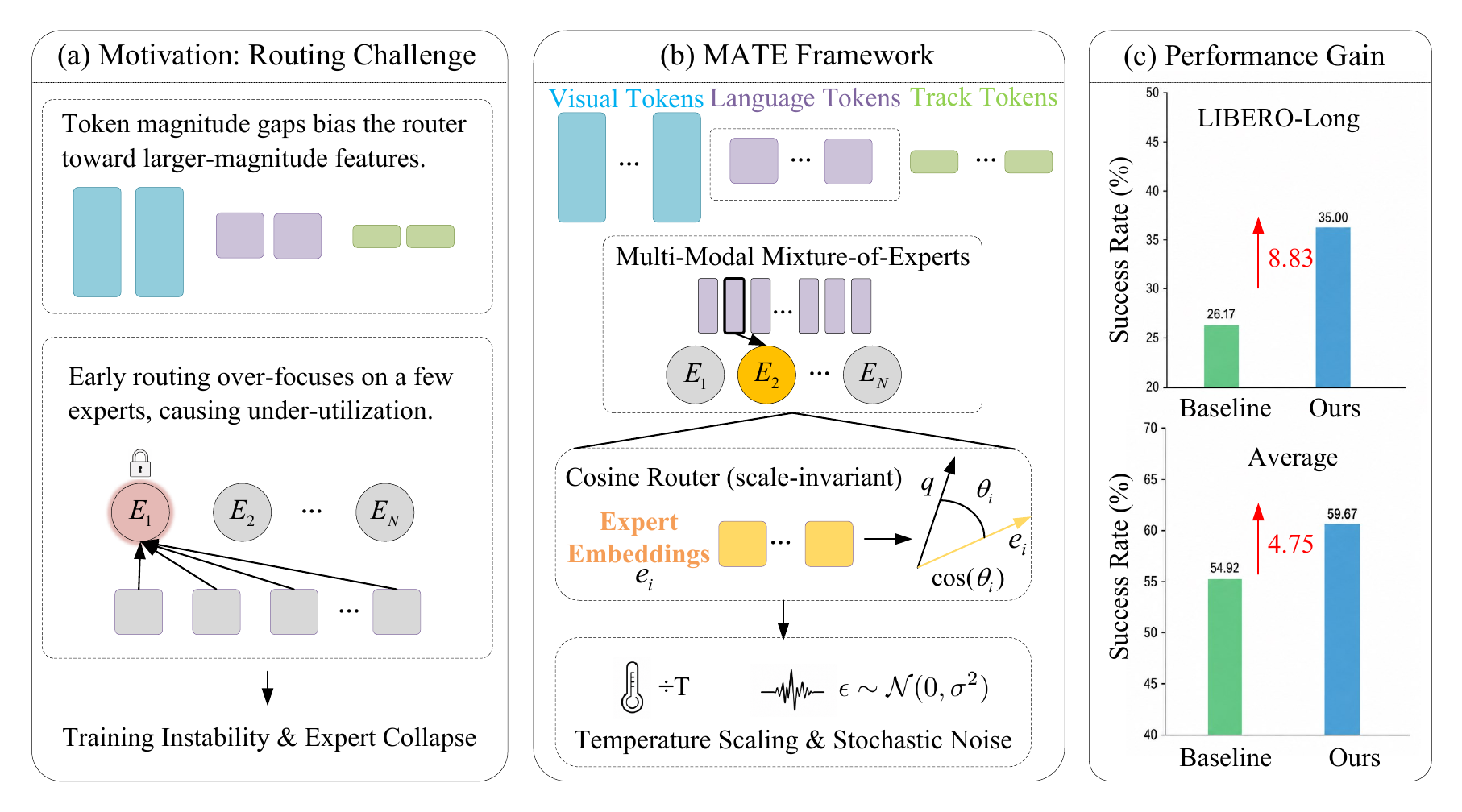}}
        \caption{
        Motivation and overview of MATE. In data-scarce MoE-based trajectory learning, heterogeneous token representations may lead to unstable training and routing behavior. MATE alleviates these issues through multi-modal expert decomposition, scale-invariant cosine routing, and temperature-noise routing stabilization.
        }
        \label{fig:Introduction}
    \end{figure}

    To address the aforementioned problems, we propose a data-efficient trajectory prediction framework, coined as \textbf{MATE}. The core insight is to improve trajectory representation capability while maintaining stable and efficient training under limited data conditions. Specifically, our method introduces three key architectural innovations. First, we introduce a \textbf{Multi-Modal Mixture-of-Experts} architecture to achieve fine-grained decoupling of heterogeneous features at the sub-token level. Second, to alleviate scale discrepancies across modalities, we design a cross-modal \textbf{Cosine Router} to reduce magnitude bias and ensure stable expert allocation. Finally, we present controlled \textbf{noise and temperature adjustment strategies} to alleviate routing polarization and premature convergence in data-scarce early training, effectively relieving expert collapse.

    To validate the effectiveness of our method, we conduct comprehensive evaluations on the LIBERO benchmark~\cite{liu_zhu_NeurIPS2023}, including LIBERO-Spatial, Object, Goal, and Long. Experimental results demonstrate that MATE achieves the best average success rate among the competitive baselines under limited-demonstration settings. Specifically, MATE improves the average success rate across all evaluated LIBERO suites by \textbf{4.75\%} over the reproduced trajectory-guided baseline. On the challenging long-horizon LIBERO-Long suite, it further achieves a larger improvement of \textbf{8.83\%}. Real-world experiments on robotic ping pong also demonstrate the immense potential of this architecture in enhancing the robustness of trajectory representations. 


\section{Related Works}
\label{sec:Related Works}

	\textbf{2.1 Multi-Modal Trajectory Policy Learning}     
    
    Traditional visuomotor policies directly map visual observations to actions~\cite{levine2016end, brohan2023rt1, zhao2023learning, li2026laser, octo2024,ddpg-per}, but suffer from compounding errors in long-horizon manipulation. Recent trajectory-guided methods alleviate this issue with motion trajectories as intermediate representations~\cite{chi2023diffusion, wen2024anypoint, florence2024dp3}, enabling policies to exploit structured motion cues. ATM~\cite{wen2024anypoint} further formulates policy learning as any-point trajectory modeling, improving data efficiency with action-free videos. However, existing trajectory-based policies usually serialize visual, language, and trajectory tokens into a shared Transformer backbone~\cite{open_x_embodiment_collaboration_2024, kim2024openvla}, where heterogeneous feature distributions may cause modality interference under limited demonstrations~\cite{yu2020gradient, li2023blip2, awadalla2023openflamingo}. Our work addresses this issue by introducing expert-based sparse representation learning into trajectory-guided policy learning.

    \textbf{2.2 Mixture-of-Experts} 

    Mixture-of-Experts architectures have achieved remarkable success in scaling model capacity through conditional computation, especially in large-scale language and vision foundation models~\cite{mu2025comprehensive, fedus2021switch, nguyen2024statistical, shazeer2017outrageously, lepikhin2020gshard, riquelme2021vmoe, moe3d, sar-jepa}. Recently, MoE has also been explored in robotics for multi-task learning, domain adaptation and trajectory prediction~\cite{yang2025tramoe, huang2025moeloco, rodriguez2026lar, mazza2026moeact}. For example, Tra-MoE~\cite{yang2025tramoe} introduces dynamic routing to pre-train trajectory models across multiple domains, showing the potential of expert specialization for robotic representation learning. However, standard token-level MoE usually requires sufficient data to learn balanced routing distributions~\cite{zoph2022stmoe, puigcerver2024softmoe, nguyen2025improving, su2024maskmoe}. In data-scarce robotic settings, routing can prematurely concentrate on a few experts, leading to load imbalance and expert collapse~\cite{yang2025tramoe, xue2024openmoe}. To address this challenge, we propose \textbf{MATE}, which redesigns the routing strategy for stable expert allocation in data-efficient robotic learning.
	

\section{Preliminaries}
\label{sec:Preliminaries}

	\textbf{3.1 Trajectory-Guided Policy Representation} 

    In robotic manipulation, policy learning aims to learn a mapping from observations to actions. Formally, given an observation $o_t$, a policy $\pi$ predicts the corresponding action $a_t=\pi(o_t)$. Recent 
    \begin{wrapfigure}{r}{0.6\textwidth}
        \centering
        \includegraphics[width=0.6\textwidth]{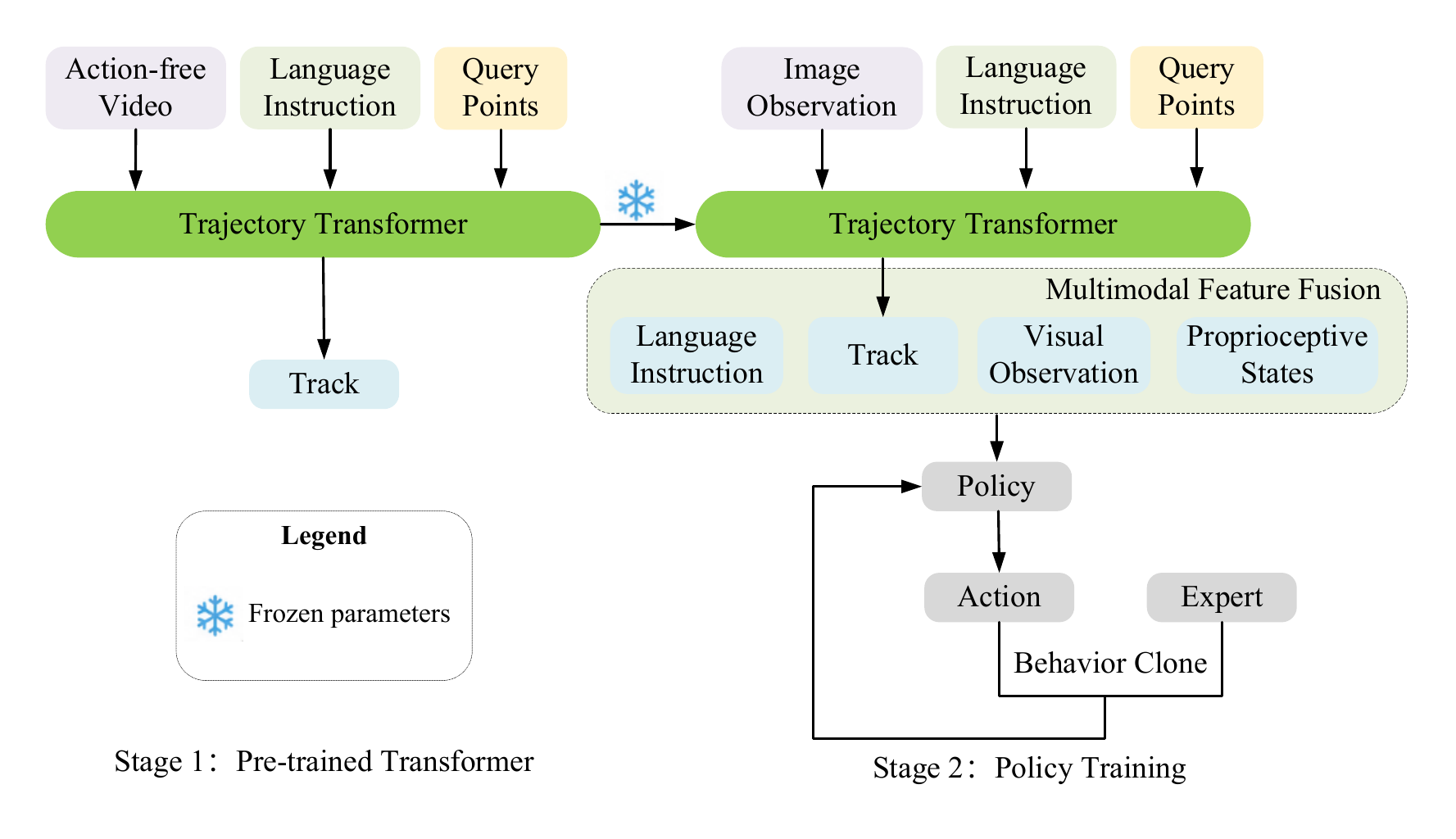}
        \caption{
        Two-stage trajectory-guided policy learning. The Trajectory Transformer is first pre-trained on action-free videos to predict future point trajectories, and then frozen to provide trajectory guidance for downstream behavior cloning.
        }
        \label{fig:ATM}
    \end{wrapfigure}
    trajectory-guided methods introduce motion trajectories as intermediate representations for policy learning. A trajectory is denoted as $\gamma=\{p_1,p_2,\dots,p_T\}$, where $p_t$ represents the robot state or end-effector pose at timestep $t$. The trajectory model predicts trajectory points conditioned on the observation $o$ and a queried timestep $t$, i.e., $p_t=f(o,t)$, allowing the policy to exploit structured motion guidance rather than relying only on direct action prediction.

    As shown in Fig.~\ref{fig:ATM}, a typical trajectory-guided framework adopts a two-stage training paradigm. In the first stage, visual observations, trajectory queries, and language instructions are encoded into heterogeneous token embeddings and integrated by a multi-modal trajectory model:
    \begin{equation}
        z_{traj} = \Phi([z_{img}, z_{query}, z_{lang}]),
    \end{equation}
    where $z_{img}$, $z_{query}$, and $z_{lang}$ denote the visual, trajectory-query, and language embeddings, respectively, and $\Phi$ represents the multi-modal trajectory modeling network.
    In the second stage, the learned trajectory representation is leveraged to facilitate downstream policy learning with action supervision. This two-stage formulation improves data efficiency by decoupling trajectory understanding from action prediction, thereby reducing the reliance on large-scale labeled demonstrations.

    \textbf{3.2 Mixture-of-Experts} 
    
\begin{figure}[htbp]
        \centering
        \includegraphics[width=1\textwidth]{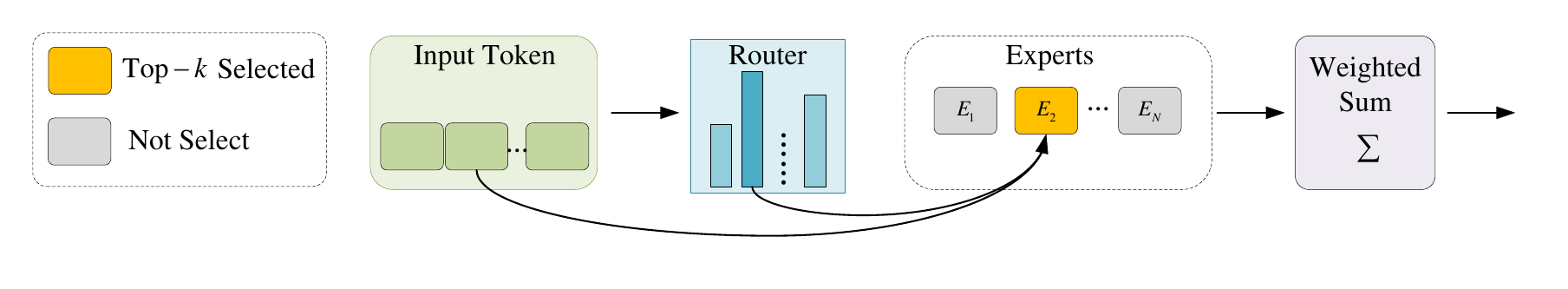}
        \caption{
        Illustration of the standard Top-$k$ Mixture-of-Experts architecture. 
        A token is routed by the router to $\mathrm{Top}\text{-}k$ selected experts. The outputs from the selected experts are aggregated to produce the final output.
        }
        \label{fig:MOE}
    \end{figure}

    MoE introduces a set of expert networks and dynamically selects a subset of them for each input token, as illustrated in Fig.~\ref{fig:MOE}. This conditional computation mechanism allows the model to scale its parameter capacity without proportionally increasing computational cost. Formally, given an input representation $x$, a router produces expert weights $g_i(x)$ and the MoE output is computed as:
    \begin{equation}
        y = \sum_{i \in \mathrm{Top}\text{-}k(g(x))} g_i(x) E_i(x),
    \end{equation}
	where $E_{i}(\cdot)$ denotes the $i$-th expert network, the routing weights satisfy $\sum_{i=1}^{N} g_i(x)=1$ and only the top-$k$ experts are activated for each token.
    By allowing different experts to specialize in different input patterns, MoE architectures significantly increase the model's representational capacity while maintaining manageable computational cost. This property has led to successful applications in large-scale models such as Switch Transformers and GLaM, where MoE enables models to scale to billions or even trillions of parameters.

\section{Method}
\label{sec:Method}

\begin{figure}[htbp]
        \centering
        \includegraphics[width=1\textwidth]{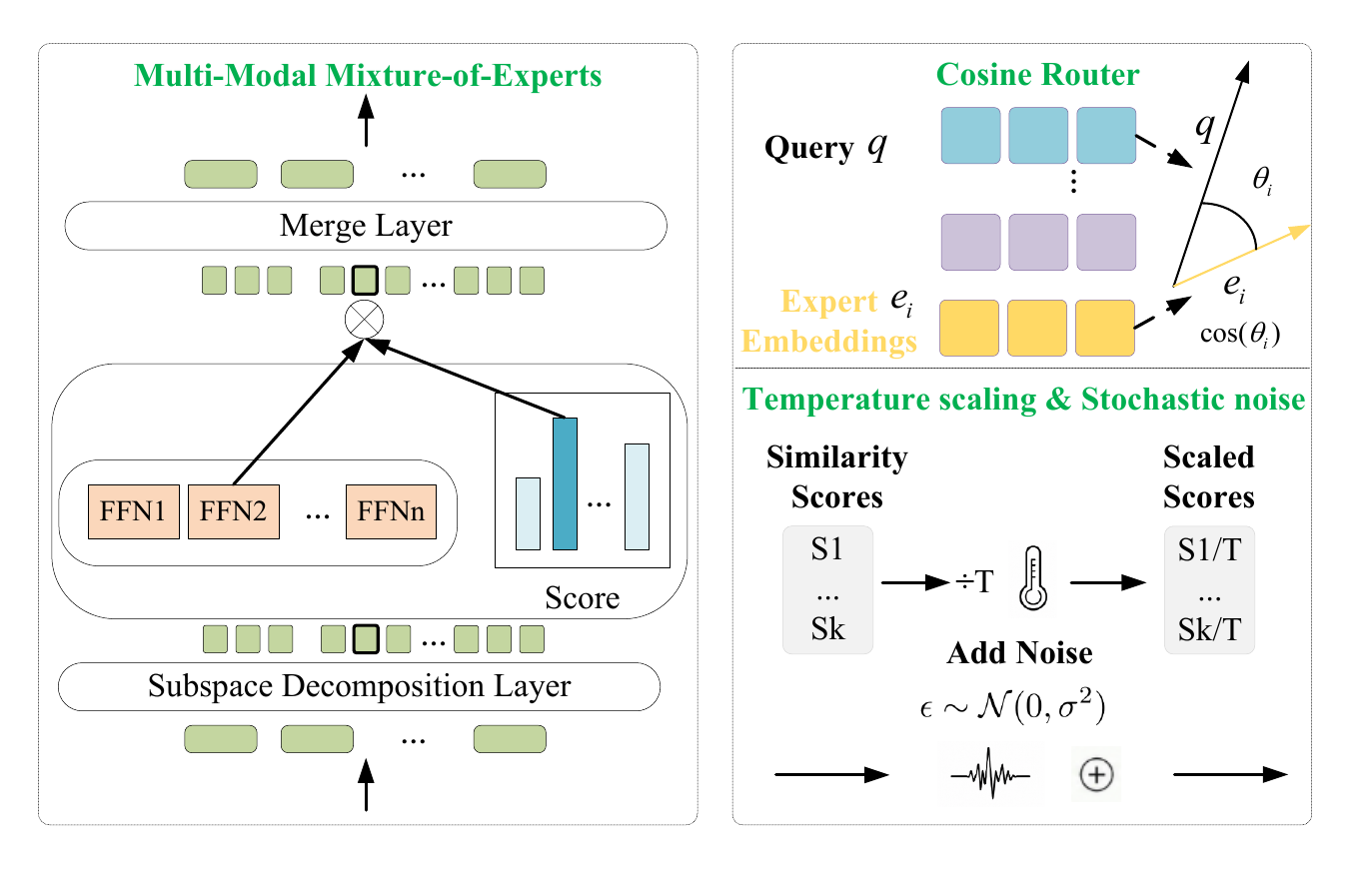}
        \caption{
        Overview of the MATE architecture. MATE consists of three key components: Multi-Modal MoE decomposes token features into subspaces for fine-grained expert routing, the cosine router computes scale-invariant expert affinities, and temperature-noise stabilization regularizes routing during training.
        }
        \label{fig:MH-MOE}
    \end{figure}

    In this section, we describe the technical design of MATE under the trajectory-guided policy learning formulation. MATE enhances the Trajectory Transformer with routing-based expert specialization to better model heterogeneous visual, language, and trajectory-query tokens under limited demonstrations. Specifically, Multi-Modal MoE enables fine-grained expert selection over decomposed token subspaces, the cosine router reduces magnitude-biased expert assignment across modalities, and temperature-noise routing stabilization alleviates premature routing collapse. The overall architecture is illustrated in Fig.~\ref{fig:MH-MOE}.

    \textbf{4.1 Multi-Modal Mixture-of-Experts}

    We first introduce the Multi-Modal MoE layer, which extends a standard FFN sublayer with multiple expert branches and routing-based selection. Unlike standard MoE layers that route each token as a whole, Multi-Modal MoE decomposes token representations into multiple feature subspaces and performs expert routing at the sub-token level.

    Formally, let the input token sequence be denoted as $\mathbf{X} \in \mathbb{R}^{l \times d}$, where $l$ represents the sequence length and $d$ denotes the embedding dimension. We first introduce a learnable projection matrix $\mathbf{W}_{mh} \in \mathbb{R}^{d \times d}$ to map the input into a diversified latent space: $\tilde{\mathbf{X}} = \mathbf{X} \mathbf{W}_{mh}.$
    To facilitate fine-grained expert routing, we partition $\tilde{\mathbf{X}}$ into $h$ distinct subspaces along the feature dimension, where $d$ is assumed to be divisible by $h$. We reshape $\tilde{\mathbf{X}}$ to isolate these subspaces, resulting in a subdivided representation $\mathbf{X}_{sub} \in \mathbb{R}^{(l \cdot h) \times (d/h)}$, where $d/h$ is the subspace dimension. Each slice $\mathbf{x}_{i,j} \in \mathbb{R}^{d/h}$ corresponds to the $j$-th subspace of the $i$-th token. This decomposition forces the subsequent routing mechanism to evaluate individual semantic facets of the token independently. Once decomposed, each subspace representation $\mathbf{x}_{i,j}$ is routed to a set of experts. We define a routing function $\mathcal{G}(\cdot)$  that computes the selection probability for a set of $E$ experts. For each subspace token, the gating output is computed as $\mathbf{P}_{i,j} = \mathcal{G}(\mathbf{x}_{i,j})$. The processed subspace feature $\mathbf{y}_{i,j}$ is obtained through a weighted sum of the active expert transformations $\mathcal{F}_e$:
    \begin{equation}
        \mathbf{y}_{i,j}=\sum_{e=1}^E(\mathbf{P}_{i,j})_e \cdot \mathcal{F}_e(\mathbf{x}_{i,j}),
    \end{equation}
    where $(\mathbf{P}_{i,j})_e$ is the routing weight for expert $e$. By performing this operation in parallel across all $h$ subspaces, we effectively increase the density of expert activation, allowing different experts to capture distinct modalities or features within the same input token without increasing the computational overhead of individual expert networks.

    To ensure that the decomposed processing benefits the overall robot policy, we reconstruct the original representation through a learnable integration step. The subspace outputs are rearranged to form the aggregated latent sequence $\mathbf{Y} \in \mathbb{R}^{l \times d}$. We then apply a refinement projection matrix $\mathbf{W}_{merge} \in \mathbb{R}^{d \times d}$ to synthesize the information gathered from diverse expert representations: $\mathbf{Z} = \mathbf{Y}\mathbf{W}_{merge}.$
    This integration mechanism ensures that the model can capture complex correlations between different sensory modalities by allowing the refined features $\mathbf{Z}$ to maintain a holistic representation for subsequent layers.

    \textbf{4.2 Scale-Invariant Routing via Cosine Similarity}
    
    Given a subspace token $\mathbf{x}_{i,j} \in \mathbb{R}^{d/h}$, a standard linear router computes its affinity score vector $\mathbf{s}_{i,j} \in \mathbb{R}^E$ across $E$ experts as $\mathbf{s}_{i,j} = \mathbf{x}_{i,j}\mathbf{W}_{gate} + \mathbf{b}_{gate}$, where $\mathbf{W}_{gate} \in \mathbb{R}^{(d/h) \times E}$ and $\mathbf{b}_{gate} \in \mathbb{R}^{E}$ are learnable parameters. To enable sparse expert activation, we apply a $\mathrm{Top}\text{-}k$ operation to $\mathbf{s}_{i,j}$, which retains the $k$ largest scores and masks the remaining entries as $-\infty$. The routing probability $\mathbf{P}_{i,j} \in \mathbb{R}^{E}$ is then obtained by applying softmax to the masked scores:
    \begin{equation}
        \mathbf{P}_{i,j} = \text{Softmax}(\text{Top-}k(\mathbf{s}_{i,j})).
        \label{eq:topk_softmax_routing}
    \end{equation}
    However, multi-modal trajectory modeling naturally involves heterogeneous inputs with different feature scales, such as visual observations, language instructions, and trajectory queries. Conventional linear routing is sensitive to these scales and can lead to biased expert utilization. To ensure robust assignment across modalities, we introduce a cosine similarity-based router. We maintain a set of learnable expert prototypes $\mathbf{C}=\{\mathbf{c}_1,\dots,\mathbf{c}_E\}$, where $\mathbf{c}_e \in \mathbb{R}^{d/h}$. For a given subspace token $\mathbf{x}_{i,j}$, we compute the similarity score for each expert $e$ by:
    \begin{equation}
        s_{i,j}^{(e)} =\frac{\mathbf{x}_{i,j}^{\top}\mathbf{c}_e}
        {\|\mathbf{x}_{i,j}\|_2 \|\mathbf{c}_e\|_2},
    \end{equation}

    $\mathbf{s}_{i,j} = [s_{i,j}^{(1)}, \dots, s_{i,j}^{(E)}]$, which is then normalized using the same $\mathrm{Top}\text{-}k$ softmax operation defined in Eq.~(\ref{eq:topk_softmax_routing}). This formulation makes the routing process scale-invariant, ensuring that expert selection is driven by semantic content rather than absolute feature magnitude, thereby stabilizing the training of the multi-modal trajectory model.

    \textbf{4.3 Routing Stabilization and Exploration}

    To further alleviate expert collapse, where the router over-utilizes a few experts and neglects others, we introduce temperature scaling and stochastic routing noise. Given the similarity scores $\mathbf{s}_{i,j}$, we first apply a learnable temperature parameter $\tau>0$: $\mathbf{s}'_{i,j} = \mathbf{s}_{i,j}/{\tau}.$
    The temperature controls the sharpness of the routing distribution: larger values encourage smoother expert assignment, while smaller values lead to more selective routing. In this way, the router can adapt its confidence during training rather than relying on a fixed routing sharpness.

    During training, we inject stochastic noise into the routing scores to encourage expert exploration:
    \begin{equation}
        \tilde{\mathbf{s}}_{i,j} = \mathbf{s}'_{i,j} + \boldsymbol{\epsilon},
        \quad \boldsymbol{\epsilon} \sim \mathcal{N}(\mathbf{0}, \sigma^2 \mathbf{I}),
    \end{equation}

    where $\sigma$ denotes the standard deviation of the Gaussian noise. This perturbation serves as a regularization strategy that prevents the router from over-fitting to biased initial assignments, thereby alleviating premature routing polarization and promoting more balanced expert utilization.


\section{Experiments}
\label{sec:result}


    Our experiments are divided into three parts. In Sec.~5.1, we compare MATE with representative trajectory-guided policy learning methods, conventional policy learning baselines, and several MoE-based variants on robotic manipulation benchmarks. In Sec.~5.2, we conduct ablation studies to examine the contribution of different routing designs, including Switch routing, cosine routing, temperature adjustment, routing noise, and multi-modal routing. In Sec.~5.3, we further analyze the routing behavior and expert utilization to explain why the proposed design improves policy learning.

    \textbf{5.1 Comparison with Baseline and MoE Variants}

    \textbf{Experimental setup.}
    We evaluate MATE on four representative LIBERO suites, including LIBERO-Spatial, LIBERO-Object, LIBERO-Goal, and LIBERO-Long, which cover spatial reasoning, object-centric manipulation, goal-conditioned control, and long-horizon sequential decision-making. For each task, all methods are trained with 10 action-labeled demonstration trajectories for policy learning and 50 action-free video demonstrations for trajectory model pre-training. The observations include RGB images from a third-person camera and a wrist camera, together with proprioceptive states, and each task is specified by a language instruction. We use success rate as the main evaluation metric.

    \textbf{Baselines.} We compare MATE with three groups of baselines. First, we include representative methods reported in ATM, including BC, R3M-finetune, VPT, and UniPi, which correspond to vanilla imitation learning, representation-based video pre-training, pseudo-action labeling, and video-generation-based planning, respectively. Second, we reproduce a trajectory-guided baseline for controlled comparison. Third, we evaluate several MoE-based variants built upon the reproduced baseline, including Switch-MoE, Cosine-MoE, Cosine-MoE+T/N, and Multi-Modal MoE. The ATM-reported results are used only as reference numbers, while the reproduced baseline and all MoE variants are implemented in our codebase.

\begin{table*}[t]
\centering
\caption{
Main results on the LIBERO benchmark.
$\dagger$ denotes results reported in prior work~\cite{wen2024anypoint}; all other results are reproduced or implemented by us.
}
\label{tab:main_results}
\resizebox{\textwidth}{!}{
\begin{tabular}{lccccc}
\toprule
Method & LIBERO-Spatial & LIBERO-Object & LIBERO-Goal & LIBERO-Long & Average \\
\midrule
BC~$\dagger$                & 39.00 $\pm$ 8.20 & 51.83 $\pm$ 13.54 & 42.50 $\pm$ 4.95 & 16.67 $\pm$ 3.66 & 37.50 \\
R3M-finetune~$\dagger$      & 49.17 $\pm$ 3.79 & 52.83 $\pm$ 2.25 & 5.33 $\pm$ 1.43 & 9.17 $\pm$ 2.66 & 29.13 \\
VPT~$\dagger$               & 37.83 $\pm$ 4.29 & 19.50 $\pm$ 0.82 & 3.33 $\pm$ 2.36 & 3.83 $\pm$ 1.65 & 16.12 \\
UniPi~$\dagger$             & 69.17 $\pm$ 3.75 & 59.83 $\pm$ 3.01 & 11.83 $\pm$ 2.02 & 5.83 $\pm$ 2.08 & 36.67 \\
\midrule
baseline model              & 64.17 $\pm$ 0.24 & 62.67 $\pm$ 4.19 & 66.67 $\pm$ 6.51 & 26.17 $\pm$ 5.2 & 54.92 \\
\textbf{MATE (Ours)} 
                            & \textbf{68.5 $\pm$ 2.45} 
                            & \textbf{69 $\pm$ 4.24} 
                            & \textbf{66.17 $\pm$ 1.93} 
                            & \textbf{35 $\pm$ 7.12} 
                            & \textbf{59.67} \\
\bottomrule
\end{tabular}
}
\end{table*}

    \textbf{Results.} The main results are shown in Table~\ref{tab:main_results}. Overall, the reproduced trajectory-guided baseline already outperforms most ATM-reported reference baselines, indicating the effectiveness of trajectory guidance for policy learning under limited demonstrations. Building on this strong baseline, MATE further achieves the best average success rate across the LIBERO suites. Specifically, MATE improves the average success rate from \textbf{54.92\%} to \textbf{59.67\%}, achieving a \textbf{4.75 \%} gain over the reproduced baseline. The improvement is especially clear on LIBERO-Object and LIBERO-Long, suggesting that MATE brings larger benefits in scenarios involving object-centric manipulation and long-horizon execution.

    However, directly introducing MoE does not necessarily lead to improvements under limited demonstrations. Our baseline + Switch-MoE variant performs slightly worse than the reproduced baseline, which is also consistent with the observation reported in Tra-MoE~\cite{yang2025tramoe}. This result indicates that the benefit of MoE does not simply come from increased expert capacity. Instead, an appropriate routing design plays a key role in making expert-based policy learning effective. We therefore further analyze different routing designs in Sec.~5.2.

    \textbf{Robotic ping pong.} We construct the real-world robotic table tennis system using Unitree Z1 Pro and the 2D gantry. A human player serves the ball and the robot is required to catch and return the ball to the opposite side. Through 100 trails, our trajectory representation can improve the ball return rate from 76\% to 85\%, demonstrating the effectiveness of our MoE-based trajectory representations.

    \textbf{5.2 Ablation on MoE Routing Designs}

    To examine how different MoE designs affect trajectory-guided policy learning, we conduct progressive ablations on the reproduced trajectory-guided baseline. We start with a Switch Transformer-style MoE module, replace its router with cosine routing, add temperature adjustment and routing noise, and finally evaluate multi-modal routing combined with the proposed stabilization strategies.

    The results are shown in Table~\ref{tab:routing_ablation}. Directly adding a standard Switch-MoE module does not improve the baseline model and performs slightly worse than the reproduced baseline model. This result is reasonable in the limited-demonstration setting, where the router may quickly concentrate on a small subset of experts before sufficient specialization is learned. The degradation is relatively small, but it indicates that naive sparse routing may not be well suited for heterogeneous policy features.

\begin{table}[t]
\centering
\caption{
Ablation study on MoE routing designs. 
All variants are built upon our reproduced baseline model implementation.
}
\label{tab:routing_ablation}
\resizebox{\columnwidth}{!}{
\begin{tabular}{lccccc}
\toprule
Method & LIBERO-Spatial & LIBERO-Object & LIBERO-Goal & LIBERO-Long & Average \\
\midrule
baseline model & 64.17 $\pm$ 0.24 & 62.67 $\pm$ 4.19 & 66.67 $\pm$ 6.51 & 26.17 $\pm$ 5.2 & 54.92 \\
baseline model + Switch-MoE & 61.5 & 60 & 65.5 & 24.5 & 52.88 \\
baseline model + Cosine-MoE & 62.5 & 64 & 61.5 & 25 & 53.25 \\
baseline model + Cosine-MoE + T/N & 62 & 62 & 68.5 & 23.5 & 54 \\
baseline model + Multi-Modal MoE & 62.5 & 59.5 & 64.5 & 24.5 & 52.75 \\
\textbf{MATE (Ours)} 
& \textbf{68.5 $\pm$ 2.45} & \textbf{69 $\pm$ 4.24} & \textbf{66.17 $\pm$ 1.93} & \textbf{35 $\pm$ 7.12} & \textbf{59.67} \\
\bottomrule
\end{tabular}
}
\end{table}
    
    Replacing the standard router with cosine routing improves the performance from 52.88\% to 53.25\%. We attribute this improvement to the scale-invariant nature of cosine similarity, which makes expert selection less sensitive to feature magnitude. Adding temperature adjustment and routing noise further improves the average success rate to 54\%, suggesting that preventing overly confident routing decisions in the early training stage is beneficial.

    The Multi-Modal MoE variant enables more flexible expert selection by decomposing token representations into multiple subspaces. However, without addressing cross-modal feature discrepancies, its performance remains limited. In contrast, our final model combines multi-modal feature decomposition, cosine-based expert assignment, and routing stabilization, achieving the best average performance. These results suggest that effective MoE-based trajectory policy learning benefits not only from expert capacity, but also from a carefully designed routing mechanism.

    \textbf{5.3 Routing Behavior Analysis}

\begin{figure}[t]
        \centering
        \begin{subfigure}{0.48\textwidth}
            \centering
            \includegraphics[width=\linewidth]{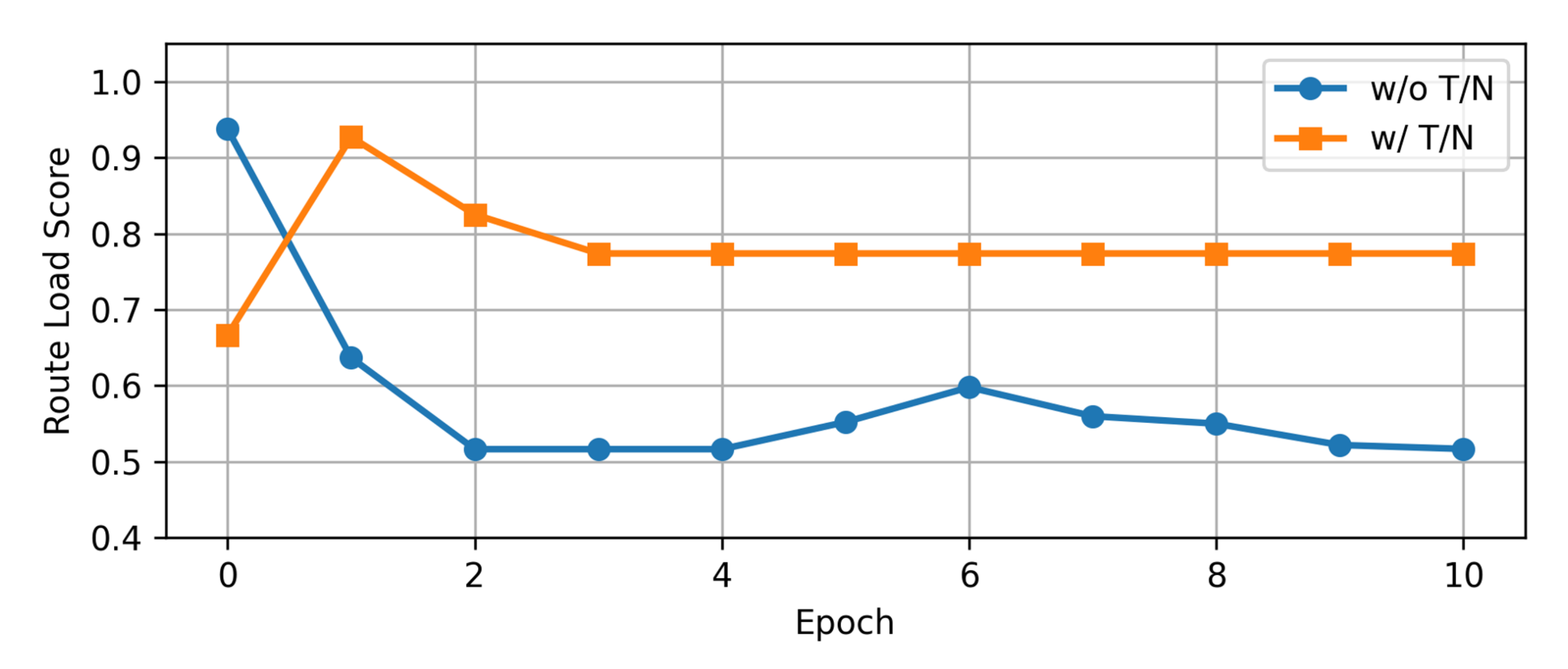}
            \caption{Route Load Score Comparison}
            \label{fig:router_load_without_tn}
        \end{subfigure}
        \hfill
        \begin{subfigure}{0.48\textwidth}
            \centering
            \includegraphics[width=\linewidth]{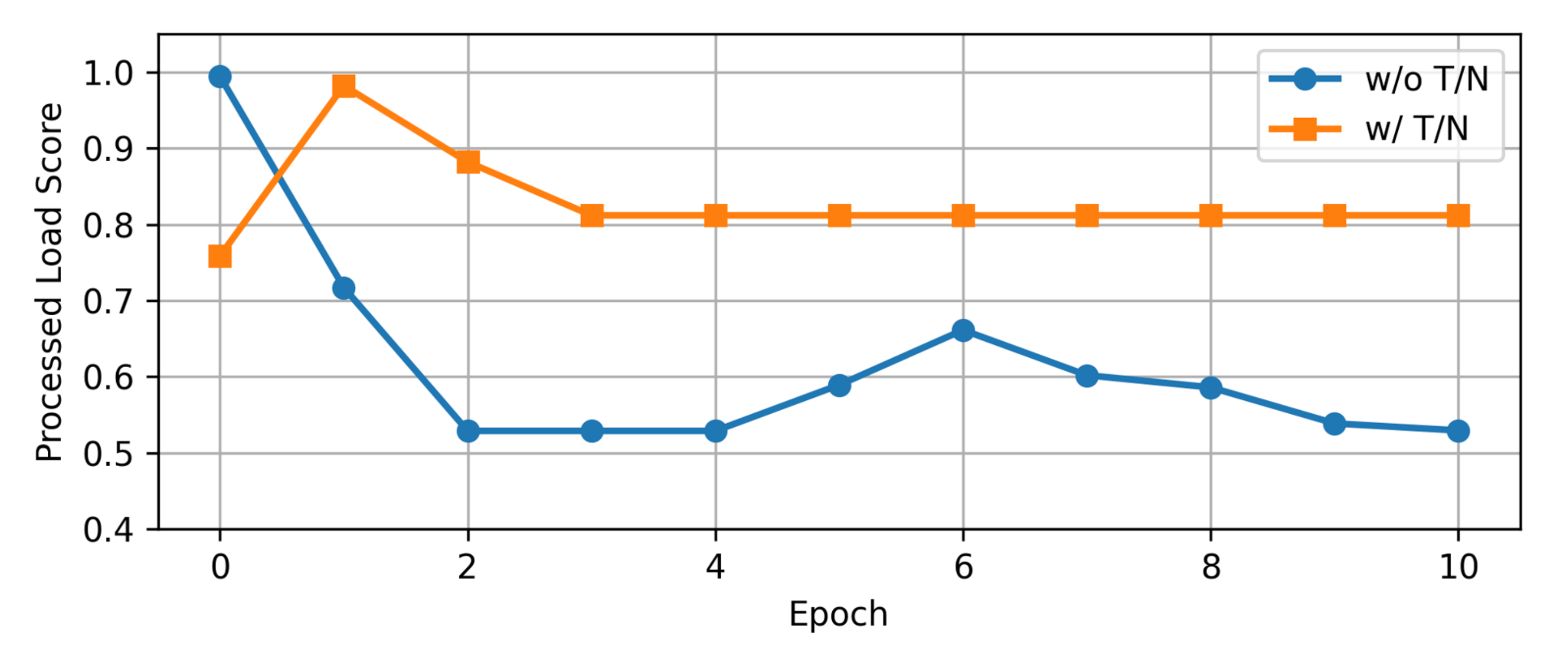}
            \caption{Processed Load Score Comparison}
            \label{fig:router_load_with_tn}
        \end{subfigure}
        \caption{
        Expert utilization comparison with and without temperature-noise routing stabilization. 
        Route load and processed load measure expert balance before and after capacity filtering, respectively. Higher scores indicate more balanced utilization.
        }
        \label{fig:router_load_compare}
    \end{figure}

    To further understand the effect of temperature-noise routing stabilization, we analyze expert utilization during training. As shown in Fig.~\ref{fig:router_load_compare}, routing without T/N stabilization exhibits lower route load and processed load scores, indicating less balanced expert utilization before and after capacity filtering. Such imbalanced expert assignment may reduce the effective capacity of MoE, which partly explains why naively introducing MoE does not necessarily bring consistent improvements over the reproduced baseline. In contrast, adding temperature scaling and stochastic routing noise consistently improves both scores, suggesting that T/N stabilization helps prevent early over-concentration on a small subset of experts and encourages more balanced expert exploration. This analysis supports the conclusion in Sec.~5.2 that stable routing design, rather than expert capacity alone, is important for effective MoE-based trajectory policy learning.

\section{Conclusion}
\label{sec:conclusion}

    We propose \textbf{MATE}, a data-efficient trajectory-guided policy learning framework for improving MoE-based trajectory modeling under limited demonstrations. 
    MATE introduces a Multi-Modal MoE architecture for fine-grained feature decoupling, a scale-invariant cosine router for robust expert assignment, and temperature-noise routing stabilization to alleviate premature routing polarization and expert collapse. 
    Experiments on LIBERO show the effectiveness of our MATE. 



    \noindent \textbf{Limitations.} Our evaluation is primarily conducted on LIBERO under limited-demonstration settings, so the application of MATE to larger-scale robotic data remains underexplored. Moreover, MATE relies on predicted trajectories as intermediate guidance, so failure cases may arise under severe visual occlusion, fast object motion, or complex contact-rich interactions. Future work will explore larger-scale training, broader real-world tasks, and validation on diverse robotic benchmarks.


\clearpage
\acknowledgments{This work was supported by National Natural Science Foundation of China (NSFC) under Grant Nos. 62376283 and 62531026; by the Fundamental and Interdisciplinary Disciplines Breakthrough Plan of the Ministry of Education of China under Grant JYB2025XDXM110; and by the Innovation Research Foundation of National University of Defense Technology under Grant JS2023-03.}


\bibliography{example}  

\end{document}